%% file: 00-Journal-Elesiver.tex
\documentclass[final,3p,times,twocolumn]{elsarticle}

\usepackage{multirow}
\usepackage{array}
\usepackage{amsthm}
\newtheorem{definition}{Definition}
\usepackage{amssymb}
\usepackage{amsmath}
\usepackage{algpseudocode}
\usepackage[ruled,vlined]{algorithm2e}
\SetKwInput{KwInput}{Input}                
\SetKwInput{KwOutput}{Output}              
\usepackage{hyperref}
\usepackage{comment}
\newcommand{\comm}[1]{}

\usepackage{lineno}

\begin{document}

\input{04-abstract}
\input{10-Introduction-final}
\input{20-Method}

\input{30-Results}
\input{50-Conclusion}

\input{91-Acknowledgement}

\bibliographystyle{ACM-Reference-Format}
\bibliography{99-references.bib}

\end{document}

%% file: 04-abstract.tex
\begin{frontmatter}

\title{Temporal cross-validation impacts multivariate time series subsequence anomaly detection evaluation\tnoteref{t1}}

\tnotetext[t1]{This manuscript has been co-authored by UT-Battelle, LLC, under contract DE-AC05-00OR22725 with the US Department of Energy (DOE). The US government retains and the publisher, by accepting the article for publication, acknowledges that the US government retains a nonexclusive, paid-up, irrevocable, worldwide license to publish or reproduce the published form of this manuscript, or allow others to do so, for US government purposes. DOE will provide public access to these results of federally sponsored research in accordance with the DOE Public Access Plan (http://energy.gov/downloads/doe-public-access-plan).}

\date{}

\author[label1]{Steven C. Hespeler\corref{cor1}}
\ead{hespelersc@ornl.gov}

\author[label1]{Pablo Moriano}
\ead{moriano@ornl.gov}

\author[label2]{Mingyan Li}
\ead{lim3@ornl.gov}

\author[label2]{Samuel C. Hollifield}
\ead{hollifieldsc@ornl.gov.}

\cortext[cor1]{Corresponding author.}

\affiliation[label1]{organization={Computer Science and Mathematics Division, Oak Ridge National Laboratory},
            city={Oak Ridge},
             state={TN 37830},
             country={USA}}

\affiliation[label2]{organization={Cyber Resilience and Intelligence Division, Oak Ridge National Laboratory},
            city={Oak Ridge},
             state={TN 37830},
             country={USA}}


\begin{abstract}
Evaluating anomaly detection in multivariate time series (MTS) requires careful consideration of temporal dependencies, particularly when detecting subsequence anomalies common in fault detection scenarios. While time series cross-validation (TSCV) techniques aim to preserve temporal ordering during model evaluation, their impact on classifier performance remains underexplored. This study systematically investigates the effect of TSCV strategy on the precision-recall characteristics of classifiers trained to detect fault-like anomalies in MTS datasets. We compare walk-forward (WF) and sliding window (SW) methods across a range of validation partition configurations and classifier types, including shallow learners and deep learning (DL) classifiers. Results show that SW consistently yields higher median AUC-PR scores and reduced fold-to-fold performance variance, particularly for deep architectures sensitive to localized temporal continuity. Furthermore, we find that classifier generalization is sensitive to the number and structure of temporal partitions, with overlapping windows preserving fault signatures more effectively at lower fold counts. A classifier-level stratified analysis reveals that certain algorithms, such as random forests (RF), maintain stable performance across validation schemes, whereas others exhibit marked sensitivity. This study demonstrates that TSCV design in benchmarking anomaly detection models on streaming time series and provide guidance for selecting evaluation strategies in temporally structured learning environments.
  \\
\end{abstract}


\end{frontmatter}

%% file: 10-Introduction-final.tex
\section{Introduction} \label{Section: Intro}
Modern interconnected systems generate vast volumes of multivariate time series (MTS) data because of continuous sensing, control, and communication activities \cite{liu2025mstvi, xiong2024siet, papastefanopoulos2023multivariate}. These datasets capture complex temporal dynamics across multiple interrelated variables, offering rich opportunities for monitoring, control, and prediction. Effectively analyzing this data for anomalies is critical for ensuring reliability, safety, and operational efficiency across such systems. However, detecting anomalies in MTS remains a significant challenge \cite{xiong2024siet, chen2025multivariate, yu2024dtaad}. This is due to the data's high dimensionality, intricate inter-variable dependencies, noise, and the presence of diverse and evolving temporal patterns. Anomalies in MTS can take various forms, e.g., point, time series, or subsequence anomalies~\cite{blazquez2021review}. In this work, we focus on subsequence anomaly detection, a practically critical yet underexplored subclass characterized by anomalous subsequences that occur following an initiating fault event. This type of anomaly is particularly important in Industry 4.0 scenarios, where early and reliable identification of fault propagation can prevent system failures, reduce downtime, and enhance operational safety.

A prominent application of subsequence anomaly detection is in fault detection, where the goal is to identify when a system transitions from normal operation to a degraded or faulty state. This task is especially important in environments where operations are continuous and time-sensitive, such as industrial automation, vehicular systems, and energy networks \cite{dong2023subsequence}. Faults often do not manifest as single, isolated anomalies but rather as sequences of abnormal behaviors that appear intermittently or evolve gradually. These complexities make intermittent fault detection in streaming multivariate data particularly challenging, requiring approaches that preserve temporal dependencies and adapt to varying fault patterns \cite{hyndman2018forecasting}. Abid et al.~\cite{abid2021review} provide a taxonomy for fault detection and diagnosis (FDD), highlighting its growing relevance beyond safety into broader domains of operational reliability and system resilience.

A key complication in fault detection within MTS lies in the temporal nature of the data. Unlike in static or spatial datasets, where the order of observations may be arbitrary, time series data inherently depends on the sequence in which data points arrive. This makes it essential not only to detect the presence of anomalies but also to identify their temporal structure and boundaries. In the case of faults, these anomalies may surface intermittently and at varying intensities, requiring models that can retain temporal continuity and detect subtle, repeated deviations. For example, a vehicle experiencing an intermittent mechanical issue may exhibit abnormal sensor readings across disjoint time windows—individually unremarkable, but collectively indicative of a persistent fault. Capturing these fault zones requires methods capable of modeling long-range dependencies and detecting evolving patterns that signal a transition from normal to abnormal states.

Recent advances in MTS anomaly detection have introduced increasingly sophisticated methods designed to capture complex interdependencies and temporal dynamics. Multivariate detection models, ranging from statistical techniques to modern deep learning (DL) architectures, have demonstrated improved capabilities in identifying outlier patterns that span multiple correlated variables \cite{zhao2024graph, xie2023anomaly, xu2021anomaly, zhao2020multivariate, audibert2020usad, li2017multivariate}. These approaches often rely on autoencoders, recurrent neural networks, and graph-based representations to learn the joint distribution of normal system behavior \cite{zhou2022contrastive, su2019robust, deng2021graph}. However, while effective in identifying individual outliers or classifying entire time series, these methods often fall short when tasked with detecting anomalous subsequences, a more complex problem that demands both high temporal sensitivity and robustness to intermittent patterns.

This limitation is particularly evident in existing benchmarking studies, which primarily focus on either point anomalies or classification of complete time series \cite{ruiz2020benchmarking, ruiz2021great}. These evaluations typically rely on static train/test splits and holdout strategies that are ill-suited to the subsequence detection setting. In the context of streaming data and real-time applications, such strategies risk overlooking the evolving nature of faults and the need for continual model adaptation. Moreover, anomalies by definition, are rare, leading to severe class imbalance and high susceptibility to false positives. While recent work has attempted to address class imbalance through oversampling and tailored loss functions \cite{lee2020fault, zhao2022t, deng2022ib}, and reduce false positives using context-aware or range-based evaluation metrics \cite{saripuddin2021random, zhou2020range}, the specific challenges of subsequent anomaly detection remain largely unaddressed.

One major gap lies in the evaluation methodology itself. Reliable detection of anomalous subsequences requires evaluation protocols that respect the temporal structure of the data and accommodate the non-stationarity of streaming environments. Conventional holdout methods, while widely used, are static by design and often incompatible with real-world MTS applications where data continuously evolves \cite{loeffel2017adaptive, gama2014survey}. Time series cross-validation (TSCV) has emerged as a promising alternative, offering temporally-aware validation procedures such as walk-forward (WF) and sliding window (SW) approaches that preserve the causal ordering of events \cite{hyndman2018forecasting}. These approaches incrementally adjust training and testing sets to simulate streaming conditions, enabling more realistic assessment of generalization and robustness. However, the implications of using TSCV strategies for MTS subsequence anomaly detection remain insufficiently explored.

In this work, we address this gap by conducting an extensive empirical evaluation of TSCV strategies in the context of intermittent fault detection using multivariate time series data. We use in-vehicle network fault data as use case. Building on the foundational work of Bifet et al.~\cite{bifet2015efficient, gomes2017adaptive, bergmeir2018note}, we adopt a temporally constrained $K$-fold framework tailored for time series. This allows us to systematically investigate how different partitioning strategies and temporal windows affect the learning dynamics and performance of classifiers. Additionally, we incorporate insights from the evolution of windowing techniques in streaming environments \cite{datar2002maintaining, bifet2023machine}, particularly their role in balancing recency and historical context through forgetting mechanisms \cite{gama2009issues}.

To ensure a comprehensive and reproducible analysis, we make the following contributions:

\begin{itemize}
\item Evaluation of TSCV impact: We quantitatively assess how different TSCV strategies affect the precision–recall characteristics of classifiers trained for intermittent fault detection in MTS.

\item Model architecture analysis: We compare the behavior of shallow models and deep learning architectures to understand whether TSCV sensitivity is architecture-dependent.

\item Classifier-level stratification: We conduct a stratified analysis of algorithm performance across TSCV conditions to identify consistent trends and vulnerabilities.

\item Partitioning sensitivity study: We examine how variations in TSCV partitioning configurations influence classifier generalization under streaming and evolving data scenarios.
\end{itemize}

To promote transparency and facilitate further research, we provide a fully documented implementation of our code base on GitHub\footnote{\url{https://github.com/TovNephesh/MTS_CAN}} and release all datasets used in this study on Zenodo.\footnote{\url{https://zenodo.org/records/12807317}}

\input{12-Related-Wroks}
\input{11-Background}

%% file: 12-Related-Wroks.tex
\section{Related works}
\subsection{Prior work closely related to the present study}
This section contextualizes our work with respect to previous work on intermittent fault detection in CAN.

Theissler~\cite{theissler2017detecting} proposed a fault detection framework that addressed the challenge of detecting both known and previously unknown fault types in automotive data. By integrating one-class and two-class classifiers into an ensemble approach, the proposed framework does not require extensive parameter tuning or deep expert knowledge, making it highly accessible and practical for industry use. They provided a comprehensive evaluation on real-world road trial data showcasing the effectiveness of the ensemble method in detecting a wide range of fault types under varying conditions, demonstrating its potential for widespread application in automotive diagnostics, predictive maintenance, and beyond. The emphasis of this study was on building an approach that can generalize across different driving scenarios and detect faults without requiring extensive expert parameter tuning. 

Zhang et al.~\cite{zhang2019tree} proposed a novel tree-based method for diagnosing intermittent connection faults within CAN, emphasizing the analysis of error event pairs and the information cost associated with different fault locations. This method was validated through laboratory tests and is shown to effectively identify the locations of intermittent connection (IC) faults in CAN networks. The approach leveraged data-link layer information to detect and localize faults without requiring physical layer information or disrupting system operations, making it a promising solution for enhancing the reliability of CAN-based systems. 

Biswal et al.~\cite{biswal2021adaptive} introduced an adaptive fault-tolerant system and optimal power allocation strategies for smart vehicles within smart cities, utilizing a CAN protocol. Authors integrated IoT modules and energy-efficient mechanisms to enhance vehicle communication systems' reliability and performance with a primary focus on autonomous vehicles. The system used a combination of the CAN protocol with probabilistic automatic repeat request (PARQ) and PARQ with fault impacts (PARQ-FI) techniques to handle message retransmissions more efficiently. The authors also incorporated an energy cost model to the system to manage power distribution effectively among the sensor nodes in the CAN network. This allowed the model to calculate the energy required for transmission and reception for optimal power usage. 

Wang et al.~\cite{wang2023diagnosis} proposed a diagnostic framework that integrates signal processing techniques with advanced network modeling to identify and localize intermittent faults in CAN networks. The diagnostic model used a combination of time-domain and frequency-domain analysis to detect anomalies and utilize the Fourier transform for frequency analysis and statistical methods for anomaly detection. The proposed method was effective in identifying the location and nature of intermittent faults in CAN networks with complex topologies. The combination of time-domain and frequency-domain analysis provided a robust approach to fault detection, reducing false positives and improving fault localization accuracy.

Wang et al.~\cite{wang2023physical} investigated the identification of intermittent open and short connection faults at the physical layer of CAN data. The study focused on developing a method to detect these types of faults based on the analysis of electrical signals. Authors mentioned that intermittent open and short circuit faults in CAN networks can lead to sporadic communication failures and are difficult to diagnose using traditional methods which was established in a previous study by the same authors in \cite{wang2022comparison}. The paper addressed the need for a physical layer diagnostic approach that can identify these types of faults early using their impact on the electrical signals in the network. The study highlighted that using the physical layer approach is successful in identifying both open and short circuit faults in the CAN network. The method provided early detection capabilities and allows for timely intervention before the faults could cause significant communication issues.

Compared to the studies mentioned above, the present paper is unique in that it focuses on benchmarking TSCV methods to optimize the performance of ML and DL models for intermittent fault detection in CAN data, which none of these studies addressed. By evaluating different TSCV strategies, this work provides a systematic approach to time series fault detection and illuminates the critical impacts of temporal validation techniques on model reliability and robustness, which are not addressed in the related studies.   

\subsection{Other prior work related to the present study}
This section we contextualize our work with respect to previous work on anomaly detection in MTS data with related or similar applications to CAN.

Cheng et al.~\cite{cheng2009detection} addressed the problem of detecting and characterizing anomalies in MTS data. In doing so, they proposed a robust algorithm designed to detect anomalies in noisy MTS by capturing the relationships among different variables. Anomalies can be challenging to define in this context because they may involve abnormal values in one or more variables, or they may correspond to unexpected changes in the relationships among variables. Authors create a method that can identify global and local anomalies while accounting for the dependencies between variables. The proposed methodology leverages a graph-based framework where a kernel matrix alignment technique forms the foundation. This alignment captures the dependency relationships among MTS variables, enabling the robust identification of anomalies by aligning predictor variables with the target variable in focus. The method performs a random walk on a graph constructed from the aligned kernel matrix. Each node represents a data point or a subsequence in the time series. Authors mention that the method is versatile enough to detect different types of anomalies like subsequence or local points.

Liang et al.~\cite{liang2021robust} introduced a robust framework for unsupervised anomaly detection in industrial MTS data, leveraging multi-time scale deep convolutional generative adversarial networks (DCGANs) equipped with a novel forgetting mechanism. The framework's capability to analyze complex, time-dependent patterns without extensive labeled data offers a significant advancement in the area. The introduction of a forgetting mechanism allows for more nuanced detection by diminishing the influence of older data, thus enhancing the model's sensitivity to new, potentially anomalous patterns. The authors used four different industrial time series datasets in their study, including Genesis Demonstrator, Satellite, Shuttle, and Gamma Polarization Detector datasets. These datasets represent a variety of MTS scenarios with both normal and anomalous data points that capture different aspects of industrial systems like sensor data, remote sensing information, and fault events. The paper demonstrated that the MTS-DCGAN framework with exponential forgetting consistently outperformed other models by achieving near-perfect F1 scores (e.g., 0.999 on the Shuttle dataset) and high MCC values (e.g., 0.997 on the Shuttle dataset). These results indicate the model's strong ability to detect anomalies, particularly in imbalanced datasets where traditional models like KNN and OCSVM struggled to achieve comparable performance. The ERR-based threshold setting further enhanced the robustness of the framework.

Ji and Lee~\cite{ji2022event} presented an anomaly detection algorithm specifically designed for hybrid control units (HCUs) in hybrid electric vehicles (HEVs). The algorithm aims to detect anomalies in the control functions of HCUs. The anomaly detection algorithm is based on a one-class SVM, which is trained using normal operation data to identify deviations indicative of faults. The study focuses on the engine clutch engagement/disengagement control and the engine start cooperative control functions. The proposed anomaly detection algorithm effectively identifies both known and unknown anomalies in HCU control functions with authors highlighting high true-positive and true-negative rates. This demonstrates that the algorithm is reliable in distinguishing between normal and abnormal behavior.

Moriano et al.~\cite{moriano2022detecting} addressed the challenge of securing CAN in modern vehicles, particularly with the influence of masquerade attacks. Authors proposed a novel approach leveraging hierarchical clustering similarity to detect masquerade attacks by analyzing how the relationships between signals change under attack conditions. The method consists of hierarchical clustering on CAN signal time series data, computing clustering similarity using CluSim~\cite{gates2019clusim}, and hypothesis testing to determine whether the observed clustering changes are statistically significant. The approached was evaluated on the ROAD dataset~\cite{verma2024comprehensive}, which contains real CAN logs with different types of attacks, including fabrication, suspension, and masquerade attacks. The hierarchical clustering approach allows the method to generalize across different vehicles, as it focuses on signal relationships rather than absolute message frequencies. 

%% file: 11-Background.tex
\section{Preliminaries}

\subsection{Anomaly detection in time-series}

An accepted taxonomy for outlier/anomaly detection in time series is the foundation of our study~\cite{blazquez2021review}. Our study is centered around MTS as the input data type, subsequences as the outlier type, and multivariate detection algorithms as the nature of the detection method. 

Following the taxonomy from \cite{blazquez2021review}, we provide a few key definitions. 

\begin{definition}[Multivariate time series]
A MTS \( \boldsymbol{X} = \{\boldsymbol{x}_t\}_{t \in T} \) is defined as an ordered set of \( m \)-dimensional vectors, where each vector \( \boldsymbol{x}_t = (\boldsymbol{x}_{1t}, \dots, \boldsymbol{x}_{mt}) \) represents observations recorded at specific time points \( t \in T \subseteq \mathbb{Z}^+ \). Note that \( \boldsymbol{x} _{it} \) denotes the observation of the \( i \)-th variable at time \( t \), for \( i=1, \ldots, m \). This representation captures the simultaneous behavior of \( m \) time-dependent variables across the time index \( t \).
\end{definition}

A subsequence $ \boldsymbol{S} = \boldsymbol{x}_p, \boldsymbol{x}_{p+1}, \ldots, \boldsymbol{x}_{p+n-1}$ of the MTS is defined as a contiguous segment of length \( n \) of the time series, where \(p\) is the starting time point of the subsequence, \(|T|\) is the total number of time points, \( n \) is the number of consecutive time points included in \( \boldsymbol{S} \), and \( p \leq |T| - n+1\). 


\subsection{Subsequence anomaly detection in MTS} \label{Section: TSCV}
A MTS anomaly detection method identifies outliers by analyzing multiple time-dependent variables together while focusing on their interactions and relationships~\cite{blazquez2021review}. Traditional techniques for detecting subsequence outliers in MTS are mainly separated into two categories: model-based and dissimilarity-based. Model-based approaches evaluate the cumulative deviation of observed vectors from their expected values over the subsequence. The deviation for each time point $i$ within the subsequence is usually measured using the Euclidean norm. The subsequence is flagged as an outlier if the cumulative deviation exceeds a predefined threshold $\tau$. Mathematically, this is expressed as:
\begin{equation}
\sum_{i=p}^{p+n-1} \left\| \boldsymbol{x}_i - \boldsymbol{\hat{x}}_i \right\| > \tau ,
\end{equation}
where $\boldsymbol{x}_i$ is the observed $m$-dimensional vector at time $i$ and $\boldsymbol{\hat{x}}_i$ is the expected $m$-dimensional vector at time $i$. The norm represents the difference between the two vectors $\boldsymbol{x}_i$ and $\boldsymbol{\hat{x}}_i$, or the difference between the observation and prediction. As dissimilarity-based techniques are rarely used in MTS, we do not discuss them here.

\subsection{Time series cross-validation (TSCV)}

Standard $k$-fold cross-validation partitions the dataset into $k$ mutually exclusive folds of approximately equal size, where the classifier is trained on \(k - 1\) folds and tested on the remaining fold~\cite{kohavi1995study}. This process is repeated $k$ times, with each fold used once as the test set. Standard CV techniques (e.g. $k$-fold~\cite{blum1999beating}, holdout~\cite{ebbes2011sense} and repeated random sub-sampling validation/Monte Carlo validation~\cite{xu2001monte, simon2007resampling}) fail to maintain temporal dependencies~\cite{cerqueira2020evaluating}. These techniques assume that data points are independent and identically distributed (i.i.d.), which is not the case with time series data where observations are inherently sequential and dependent on prior values. Traditional CV approaches can lead to data leakage because it might train the model using future data points while testing on past data. 

Due to temporal dependencies, standard $k$-fold CV is not a suitable method for evaluating a time-series dataset due to temporal dependencies present in the data (unfathomable to utilize future data to train and historic data to test) \cite{roberts2017cross, jiang2017markov, ramos2016procedure, sheridan2013time}. More specifically, $k$-fold CV is not effective for periodicity, overlapping, or correlation of time series data \cite{jiang2017markov}. That is, $k$-fold CV needs to deal with the observations sequentially upon arrival to the system~\cite{moriano2024benchmarking, jiang2017markov}. 


TSCV methods in this study can be categorize as either holdout or prequential~\cite{gama2014survey}. The holdout method segments a data stream into distinct training and testing sets, where the model is trained on a fixed historical dataset and evaluated on a separate test set while maintaining temporal dependencies. Alternatively, prequential methods update the model incrementally as new streaming data becomes available. In our study, we quantify the effect of two prequential-based techniques: (1) WF and (2) SW. On one hand, WF validation incrementally expands the training dataset to include observations as they enter the system into the testing data, allowing the model to adapt to new information iteratively~\cite{olorunnimbe2023deep, hyndman2018forecasting}. Thus, each iteration uses the updated training set to predict subsequent observations, ensuring temporal integrity and reflecting the evolving nature of the data stream. On the other hand,  SW dynamically partition the data stream into manageable chunks based on temporal or count-based criteria.  SW maintain a fixed-size window of recent observations, shifting incrementally to accommodate new data while discarding the oldest, effectively capturing temporal dynamics~\cite{patroumpas2006window, gama2013evaluating, gama2014survey, verwiebe2023survey}. Prequential-based techniques like WF and SW are particularly well-suited for online learning in dynamical environments and preserve temporal dependencies that are crucial for robust evaluation~\cite{hyndman2018forecasting, gama2014survey, gama2009issues, gama2013evaluating, bifet2015efficient}.

\subsubsection{Training/testing subsequences}
Using the definition from Kohavi~\cite{kohavi1995study}, for each fold $k$ we define two sets. The training set, $\boldsymbol{S}_{train}^{k}$, contains the set of observations used to train the model. The testing set, $\boldsymbol{S}_{test}^{k}$, includes a set of observations used to evaluate the model. The distinction between these sets ensures that the model is trained on historical data while being evaluated on unseen, future data. This division varies depending on the specific TSCV strategy employed. Mathematical, it can be written as:  

\begin{equation}
    \boldsymbol{S}_{train}^{k} = \{\boldsymbol{x}_t, y_t | t \in T_{train}^{k}\}
\end{equation}

\begin{equation}
    \boldsymbol{S}_{test}^{k} = \{\boldsymbol{x}_t, y_t | t \in T_{test}^{k}\} ,
\end{equation}

\noindent where, $y_t$ represents the corresponding label of the $\boldsymbol{x}_t$ observations and $T_{train}^{k}$ and $T_{test}^{k}$ are a set of time indices forming the train and test sets, respectively. Note that, in a specific fold $k$, $T_{train}^{k}$ contains the time steps up to a certain time index and is used in the training phase and the $T_{test}^{k}$ contains the time steps that the model has not seen during training and is used to evaluate the model's performance. 

\subsubsection{Learning and prediction}
The training phase involves fitting a classifier, $\mathcal{M}_{k}$, to the training subsequence $\boldsymbol{S}_{train}^{k}$. This process optimizes the model parameters, $\theta_k$, that $\mathcal{M}_{k}$ learns from the training subsequence $\boldsymbol{S}_{train}^{k}$ by optimizing its parameters $\theta_k$ to capture the underlying patters and relationships in the data. Mathematically, the training process is represented as:

\begin{equation}
    \mathcal{M}^*_k: \mathcal{M}_k \left(\boldsymbol{S}_{train}^{k}, \theta_k\right).
\end{equation}

Once trained, $\mathcal{M}^*_k$, where $\theta^*_k$ is the optimized parameter set from the $k$-th fold, it is used to generate predictions for $\boldsymbol{S}_{test}^{k}$. The predictions, $\hat{y}_{test}^{k}$, are computed as:

\begin{equation}
    \hat{y}_{test}^{k} = \mathcal{M}^*_{k}\left(\boldsymbol{S}_{test}^{k}, \theta^*_k\right),
\end{equation}

\noindent where, $\hat{y}_{test}^{k}$ represents the predicted labels for the testing subsequence. Classification evaluation is performed for each $k$-fold and is assessed using the predicted outcomes, $\hat{y}_{test}^{k}$, against the ground-truth labels, $y_{test}^{k}$. 


\subsection{CAN protocol}
CAN, introduced by Bosch \cite{specification1991bosch}, is a serial communication protocol that effectively facilitates distributed real-time communication among electronic control units (ECUs). ECUs are fundamental to modern vehicles and are responsible for managing everything from engine operations to safety systems like anti-lock brakes and airbag deployment. Each ECU is specialized for controlling specific subsystems, making communication between these units vital for vehicle performance and safety.

CAN is applied across a spectrum from high-speed networks to economical multiplex wiring solutions. Within the realm of automotive electronics, components such as engine control units, sensors, and anti-skid systems are interconnected through CAN, supporting data transmission speeds of up to 1 Mbit/s.

\begin{figure*}[!t]
    \centering
    \includegraphics[width=\linewidth]{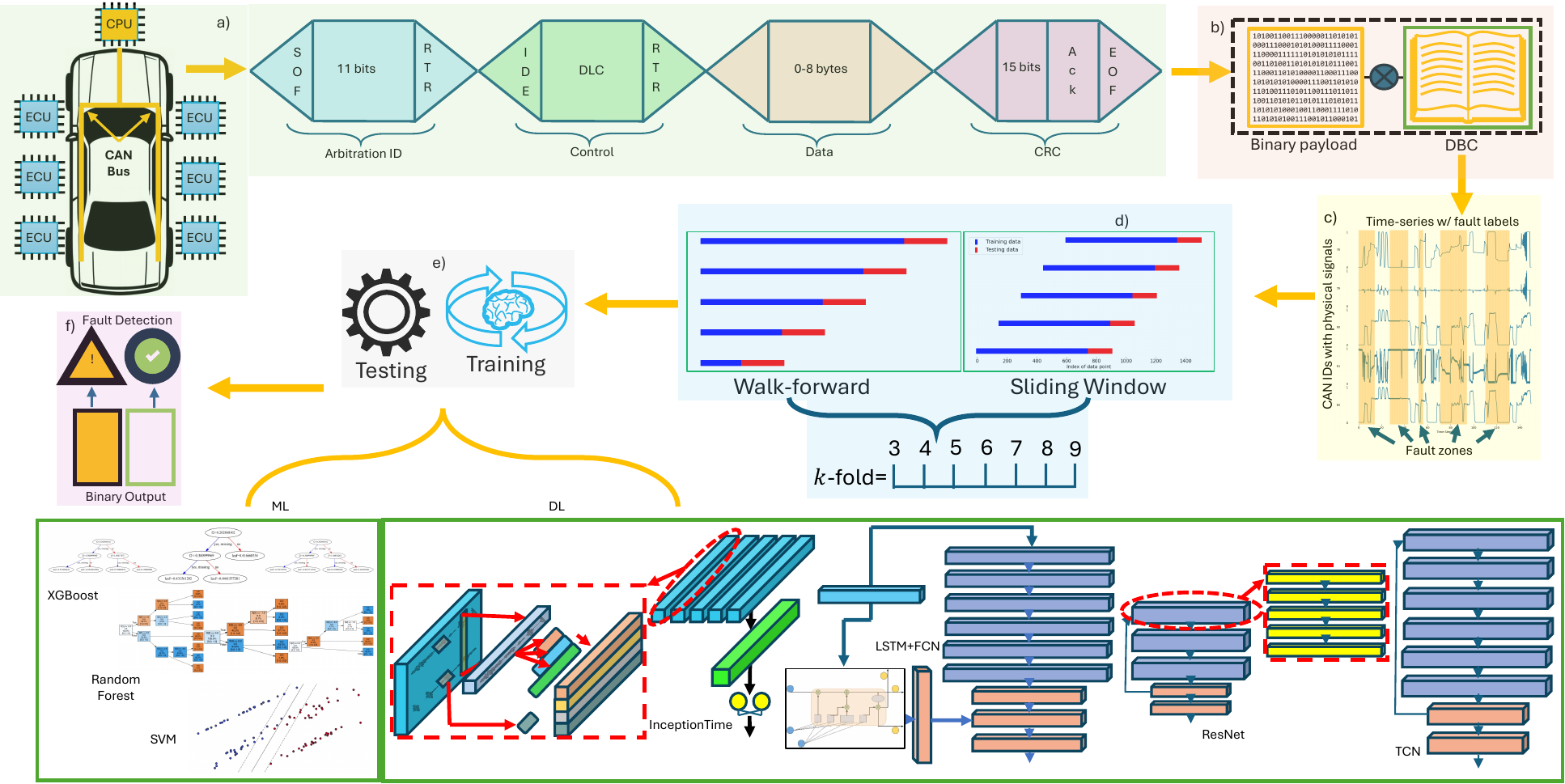}
    \caption{Design diagram illustrating the experimental design process including: a) CAN data acquisition with main fields from left to right: Arbitration ID, control ID, and CRC fields, b) binary payload and DBC used for reverse engineering MTS, c) time series data with fault zones, d) application of TSCV methods with $k$-fold range (3-9), e) training and testing of models, and f) fault detection using binary classification.}
    \label{fig:design_dia}
\end{figure*}

A typical CAN data frame is illustrated in Fig. \ref{fig:design_dia} (a), it comprises of various segments, notably the 11-bit arbitration ID and the 64-bit data field or payload, which are crucial for the discussions in this paper. The arbitration ID represents a unique identifier of CAN frames transmission over the bus. Thus, the ID helps to distinguish between frames and determine message priority. Here, lower ID values means higher message priority, or frames that need to be sent with lower delay. On the other hand, the payload contains specific pieces of information known as ``signals" spanning across multiple bits in the payload. That means that the payload encodes sensor readings or status information, which is converted into MTS data.

%% file: 20-Method.tex
\section{Methodology}
This section delineates the scientific methodology employed to conduct our study on the impact that TSCV methods have on the detection of subsequence anomalies. First, we define the three TSCV methods that we benchmarked \cite{ramos2016procedure, hyndman2018forecasting} (Section \ref{Section: TSCV methods}). Next, we detail the shallow and DL classifiers used (Section \ref{Section: TSCV_algos}). Then, we discuss the evaluation metrics for comparing performance of the classifiers (Section \ref{Section: mod_inference}). Finally, we provide details on the data collection and preparation process of the CAN data (Section \ref{Section: Exp_setup}).  


\input{21-Experimental-Setup-final}



%% file: 21-Experimental-Setup-final.tex
\subsection{TSCV methods} \label{Section: TSCV methods}

We studied two different prequential approaches: WF and SW. Next, we highlight details of each method. 
\begin{enumerate}
    \item \textbf{WF approach over MTS stream}: In this method, the data is divided sequentially, starting with an initial training sample \(\boldsymbol{x}_1\) that includes the earliest data points in the series \cite{olorunnimbe2023deep, hyndman2018forecasting}. The model is trained on this initial subsequence and tested on the next sequential data point or a small block of data points immediately following the training subsequence. After each iteration, the training subsequence is expanded to include the previously tested data points, and the block of data is used for testing as observations arrive.
    
    \begin{definition}[WF]
    Initialize the training subsequence and expand the window by adding more observations over time. For each k-fold, where each fold \(k \in \{1, \dots, K\} \), the training subsequence is: 
    
    \[ \boldsymbol{S}_{train}^{k} = \{\boldsymbol{x}_1, \boldsymbol{x}_2, \dots, \boldsymbol{x}_{\omega + (k-1)\delta}\}. \] 
    
    The test subsequence immediately follows the training subsequence and is defined as: 
    
    \[\boldsymbol{S}_{test}^{k} = \{\boldsymbol{x}_{\omega + (k-1)\delta+1}, \dots, \boldsymbol{x}_{\omega + k\delta}\}, \] 
    
    where, \(\boldsymbol{x}_1\) is the initial training sample in the training subsequnce, \(t\) is the current time, \(\omega\) is the window length, and \(\delta\) is the offset. The window length of the training subsequence defines the number of observations included in each training window. The offset \(\delta\) represents the number of observations added to the training subsequnce after each iteration. The window grows at \(\omega + (k-1)\delta\) at each fold. As the window advances forward by a fixed offset \(\delta\), the classifier is trained on incremental data ensuring sequential evaluation over time.

    \end{definition}
    
    \item \textbf{SW over MTS streams:} 
    We employ a sample-based SW approach, defined by Patroumpas, et al.~\cite{patroumpas2006window}, where the MTS stream is partitioned into a fixed-size training window and a fixed-size testing window. The fixed-bound windows are adjusted to fit each $K$-fold experiment. The training window slides as new observations enter the system, disregarding old observations while the testing window slides forward incrementally.

\begin{definition}[Sample-based SW]
A time-based SW is defined over a subsequence of the MTS data stream \( \boldsymbol{S} \). This condition specifies which observations of the stream are included in the window at time \( t \). The window returns a finite subsequence of observations. The state of the window evolves over time by adding new observations and expiring old ones. For each $K$-fold, where each fold \(k \in \{1, \dots, K\} \), the training subsequence at the $k^{th}$ fold is: 
\[ \boldsymbol{S}_{train}^{k} = \{\boldsymbol{x}_{1+(k-1)\delta}, \boldsymbol{x}_{2+(k-1)\delta}, \dots, \boldsymbol{x}_{\omega+(k-1)\delta}\} ,\] 

and the test subsequence immediately follows the training subsequence and is defined as: 

\[\boldsymbol{S}_{test}^{k} = \{\boldsymbol{x}_{\omega+(k-1)\delta+1}, \dots, \boldsymbol{x}_{\omega+k\delta}\}.\] 


\end{definition}


\end{enumerate}

\input{24-Machine-Learning-final}

\subsection{Data collection} \label{Section: Exp_setup} 

\begin{figure}[!b]
    \centering
    \includegraphics[width=\linewidth]{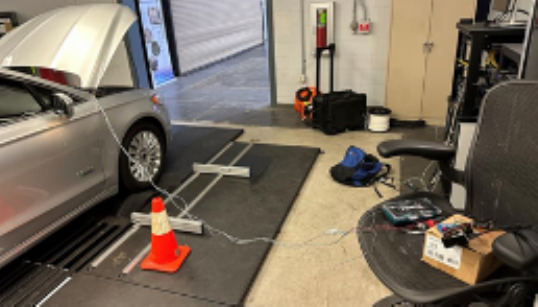}
    \caption{Experimental setup for capturing CAN data from a vehicle. The setup includes a car positioned on a testing platform, a laptop, and additional diagnostic equipment are arranged nearby to monitor and record the CAN data transmitted by the vehicle's ECUs during the experiment.}
    \label{fig:CAN_setup}
\end{figure}

A 2013 Ford Fusion Hybrid EnergiMax was positioned on a dynamometer for stationary data collection. A Raspberry Pi was connected to the CAN bus via the OBD-II port and a 4-channel relay was used to manipulate the (on/off) state of the (1) power injector and (2) spark plug following a similar experimental setup from Thesissler~\cite{theissler2017detecting}. Fig. \ref{fig:CAN_setup} visualizes the experimental setup for collecting CAN data.

\subsubsection{Fault injection}
All data was collected at the National Transpiration Research Center (NTRC) at Oak Ridge National Laboratory (ORNL) facilities. To simulate faults for fault evaluation, faults were deliberately introduced into the vehicle's CAN network. This was performed by manipulating vehicle components by toggling the relay to simulate faults and log the timestamps of these events. To simulate a faulty power injector and spark plug, the power injector/spark plug was triggered on/off through the relay system. The vehicle was stationary with frequent gas pedal presses to engage the combustion engine. Faults were intermittently injected 6 times for the power injector and 5 times for the spark plug, with all fault zones lasting for various duration's ranging from 4 seconds to 25 seconds. 

Time stamps in the experiment were collected using the ``timestamp" attribute from CAN messages received by the Raspberry Pi. When a fault was introduced or resolved, the code recorded the current time stamp from the CAN message to accurately record time of the event. By doing this, we record precise tracking of fault duration and provide a robust dataset for evaluating fault detection models. 


\subsubsection{Data preparation} \label{subsection: data_prep}
In our methodology, we pre-process the data for temporal subsequence classification across eight distinct ML classifiers. Initially, we segregate predictor variables and the target variable ``Fault Status" from our dataset. For classifiers requiring three-dimensional input, we reshape the standardized data to include a `channels' dimension, accommodating the requirements of convolutional neural network-based architectures. This pre-processing pipeline standardizes the input data format across all classifiers, ensuring consistency in data presentation and facilitating comparative analysis of model performance. We begin the pre-process with the transformation of raw CAN captures from binary payload into structured MTS formats (see Fig. \ref{fig:design_dia}(c)). In doing so we use CAN-D~\cite{verma2021can}, to reverse engineer CAN signals, yielding a detailed representation of the vehicle's operational dynamics. To ensure uniform temporal resolution across all signals, we resampled the time series at a common rate of 100 Hz, reflecting the observation that most IDs do not transmit data at frequencies exceeding this threshold. This approach harmonizes the differing native sampling rates associated with each ID by aligning them on a consistent timeline.

To ensure the comparability and analysis of the signals, the derived MTS data undergo min-max normalization to ensure values between $[0,1]$. This step adjusts the data to a common scale without distorting differences in the ranges of values, making it more suitable for subsequent analytical processes. Particularly, the normalization process facilitates the accurate computation of correlation matrices.

Furthermore, the data preparation stage addresses the temporal aspect of fault detection by aligning fault state information with the time series data. This alignment is achieved by interpolating fault states to match the time stamps of the CAN captures, thereby ensuring that each point in the time series is associated with an accurate fault status. As of today, CAN-D is still a state-of-the-art reverse engineering framework for CAN data \cite{buscemi2023survey}.

\subsubsection{Fault datasets}
We produced two fault datasets per each of the faults introduced (i.e., power injector and spark plug). Both datasets are organized by CAN ID numbers containing individual signals from ECU. The power injector dataset contains 328 time series and 2,490 time stamps. Classes are roughly represented equally, with 1,318 total observations under normal status and 1,172 observations under fault status. The spark plug fault dataset contains 311 time series, 1,500 time stamps, 719 observations under normal status, and 781 observations under fault status. The MTS datasets are available to the public for reproducibility purposes on Zenodo.\footnote{\url{https://zenodo.org/records/12807317}}



%% file: 24-Machine-Learning-final.tex
\subsection{ML classifiers} \label{Section: TSCV_algos}
Binary classifiers were trained using time series features from labeled subsequences. Selected classifiers were hand-picked from state-of-the-art literature due to their popularity and high performance capabilities \cite{ismail2019deep, ruiz2021great, theissler2017detecting}. We detail them below. For the subsequent methods, we aim to minimize error via the loss function. Thus, for the binary MTS evaluation task, we use log loss (binary cross-entropy).


Below we describe the state-of-the-art classifiers and group them by shallow and deep architecture in chronological order. 

\subsubsection{Shallow classifiers}
\noindent \emph{Support Vector Machine (SVM):}
SVMs are characterized by three core components: the optimization problem, the decision function, and kernel methods \cite{cortes1995support}. SVMs have established themselves as essential tools in machine learning, excelling in both classification and regression tasks. 

The central goal of SVMs is to identify a hyperplane that best separates classes within the feature space. In cases of linearly separable data, SVMs seek a linear hyperplane that maximizes the margin, or the distance between the hyperplane and the closest data points from any class. This margin maximization is critical as it affects the SVM's capacity and generalization error. Mathematically, for binary classification with training data labeled as $y\in[+1,-1]$, SVM finds parameters $w=[w_1w_2...w_n]^T$ with a $b$ discriminant, the hyperplane is defined by the equation:

\begin{equation}
    d(x, w, b) = x^Tx+b=\sum_{i=1}^nw_ix_i+b .
\end{equation}

In cases where the data are not linearly separable, SVM employs kernel functions to map the input features into a higher-dimensional space where a linear separation is possible. The choice of kernel function is pivotal and defines the feature space in which the linear classifier operates. The polynomial and Gaussian Radial Basis Function (RBF) are common choices, with the RBF defined as:

\begin{equation}
    K(x,x')=exp(-\gamma||x-x'||^2)) ,
\end{equation}

\noindent where $\gamma$ is a parameter that determines the spread of the kernel. The SVM formulation can be solved as a convex optimization problem, specifically a quadratic programming problem. This optimization ensures that the solution is global and unique. The dual form of the SVM allows for the inclusion of the kernel function having a solution by the saddle point of the Lagrange functional and is given by:

\begin{equation}
    L(w, b, \alpha)=\frac{1}{2}w^Tw-\sum_{i=1}^l\alpha_i{y_i[x^Tx_i+b]-1} ,
\end{equation}

\noindent subject to $\sum_{i=1}^ly_i\alpha_i=0$ and $0\leq \alpha_i\leq C$; where \(\alpha_i\) are the Lagrange multipliers, $C$ is the penalty parameter, and $l$ is the number of training samples.



\noindent \emph{Random Forest (RF):}
As noted by Petković et al.~\cite{petkovic2018improving}, in examining ensemble methods we recognize the advantages of ensemble trees for its durability and interpretability. Originated by Breiman \cite{breiman2001random}, RF is celebrated for its ability to reduce variance primarily through bootstrap aggregation, commonly known as bagging. This technique involves generating multiple decision trees from bootstrap samples of the dataset, which collectively vote for the most likely class, thus boosting classifier stability without significantly increasing variance \cite{hastie2009random}.

The RF algorithm incorporates randomness in selecting input variables by randomly picking a subset of $m$ variables from the total set of $pl$ variables, with $m = \sqrt{l}$. During the training of each tree on these subsets, the classifier executes random splits on the predictors, choosing the optimal split to divide the data. This method avoids pruning and relies instead on standard tree stopping criteria. Mathematically, the RF predictor for classification tasks is defined by: 

\begin{equation}
\hat{C}^B_{rf}(x) = \text{majority vote} {(\hat{C}_b(x))}^B_1,
\end{equation}

\noindent where $B$ denotes the number of trees, and $x$ is the new data point. A key objective in our project was the identification of the most influential features during the building process, which are critical for predicting part defects with high accuracy.

\noindent \emph{Extreme gradient boosting (XGBoost):}
The XGBoost represents an advanced iteration of the traditional boosting framework, delivering scalable predictive data analysis solutions. Developed by Chen and Guestrin~\cite{chen2016xgboost}, XGBoost incorporates a design optimized for sparse data and employs a weighted quantile sketch algorithm to facilitate approximate tree learning. The algorithm's scalability is achieved through its sparsity-aware structure and the innovative use of the weighted quantile sketch, enabling it to efficiently manage large-scale and weighted datasets. The algorithm's rank function is given by:

\begin{equation}
r_k(z) = \frac{1}{\sum_{(x,h)\in\mathcal{D}k}h} \sum{(x,h)\in\mathcal{D}_k,x<z}h,
\end{equation}

\noindent where $r_k$ is the rank function, $\mathcal{D}_k$ is a multi-set dataset $\mathcal{D}_k=\{(x_{1k},h_1),(x_{2k},h_2)...(x_{nk},h_n)\}$ with $x$ predictors, $h_i$ are weights, $k$ is a feature value, and, $\in$ is an approximation factor. The XGBoost algorithm employs a novel distributed weighted quantile sketch algorithm for handling weighted data. This allows for the creation of efficient, approximate splits on continuous features, circumventing the computational intensity of exact greedy algorithms.

\subsubsection{DL classifiers}

\noindent \emph{Residual network (ResNet):}
ResNet \cite{he2016deep} introduced a novel concept of shortcut connections, or skip connections, that bypass one or more layers in the network \cite{wang2017time}. These connections facilitate the direct propagation of gradients to earlier layers, significantly mitigating the vanishing gradient problem that is prevalent in deep neural networks (DNN). This feature is especially crucial for time series data, which often requires the capture of long-term dependencies through deep network structures. The core of the ResNet architecture is its residual blocks, where the input to a block is directly added to its output, forming the final output of the block. Mathematically, this can be expressed as:

\begin{equation}
    y = F(x, \boldsymbol{W_i}) +x ,
\end{equation}

\noindent where $x$ represents the input to the block and $y$ is the output. $F(x,{W_i})$ denotes the residual function learned by the layers within the block. The weights ${W_i}$ are optimized during training. 

For our time series classification task, we adapted the ResNet architecture to efficiently process 1D temporal data. Each residual block in our classifier comprises convolutional layers with filters of varying sizes, designed to capture patterns over different temporal scales. The classifier architecture is sequentially composed of three residual blocks, each followed by batch normalization and ReLU activation for non-linearity. The final feature maps are aggregated using global average pooling, leading to a dense layer with a sigmoid activation function to suit the binary classification task.

\noindent \emph{Temporal convolutional networks (TCN):}
TCN offer a powerful approach to sequence modeling by leveraging convolutions across time \cite{tang2021channel}. The architecture consists of convolutional layers that capture temporal evolution, 1D pooling or upsampling for efficient long-duration computation, and channel-wise normalization for stabilization. The encoder-decoder framework in TCNs uses filters to process input signals over time, capturing patterns across sequences with variable lengths. The classifier predicts class probabilities for each frame, employing techniques like Leaky ReLU for activation and softmax for final classification. TCNs are versatile, allowing modifications such as skip connections and various convolutional layer patterns, to improve performance across different datasets.

\noindent \emph{Long short-term memory + fully convolutional network (LSTM + FCN):}
Long short-term memory (LSTM) networks are an advanced type of recurrent neural networks (RNNs) developed to address the issue of vanishing gradients \cite{hochreiter1997long}. This method updates the weights influenced by the loss function through standard gradient search. The main components of the LSTM memory cell include a cell state $(c_t)$, input $(x_t)$, and output $(y_t)$, all at time $t$, $t=1,2,...,T$, $x_t \in R^m$ \cite{fuqua2022commodity}. Four gate operations dictate the flow of information within the LSTM memory cell at time $t$, including the forget gate control $(f_t)$, gate control $(g_t)$, input gate control $(i_t)$, and output gate control $(o_t)$. $\textbf{W}_{xi}, \textbf{W}_{xf}, \textbf{W}_{xg}, \textbf{W}_{xo}$ are weight matrices of each layer for input $x_{(t)}$. $\textbf{W}_{hi}, \textbf{W}_{hf}, \textbf{W}_{hg}, \textbf{W}_{ho}$ are weight matrices of each layer with respect to the previous hidden state $h_{t-1}$. Finally, $b_i, {b_f}, b_g, and \ b_o$ are bias terms and {$\sigma$ is the sigmoid activation for a gate at time $t$} 

As epochs progresses, the forward pass consists of:

\begin{equation}
        \begin{aligned}
            f_t &= \sigma(\textbf{W}_{xf} \cdot x_t + \textbf{W}_{hf} \cdot h_{t-1} + b_f)  \\
            i_t &= \sigma(\textbf{W}_{xi} \cdot x_t + \textbf{W}_{hi} \cdot h_{t-1} + b_i)  \\
            o_t &= \sigma(W_{xo} \cdot x_t + \textbf{W}_{ho} \cdot h_{t-1} + b_o) \\
            g_t &= tanh(\textbf{W}_{xg} \cdot x_t + \textbf{W}_{hg} \cdot h_{t-1} + b_g) \\
            c_t &= f_t \odot c_{t-1} + i_t \odot g_t  \\
            h_t &= o_t \odot tanh(c_t) .
        \end{aligned}
\label{eq:lstm_equations}
\end{equation}

The fully convolutional network (FCN) utilizes convolutional layers, followed by batch normalization and ReLU activation, to efficiently extract features. The FCN structure incorporates a series of 1-D convolutional kernels of diverse sizes, enabling the capture of temporal patterns across various scales. This process ultimately leads to a dense layer with softmax activation for classification.


Wang et al. proposed a strong method for TSC through the use of a FCN \cite{wang2017time}. The FCN is utilized as a feature extractor. The design of the block begins with a convolutional layer proceeded by a batch normalization layer, and a Rectified Linear Unit (ReLU) activation layer. Authors utilize a convolution operation performed by three 1-D kernels with the sizes {8, 5, 3} with no striding used.  The final layer (dense layer) uses a softmax output. The convolution block is:

\begin{equation}
    y=a(\boldsymbol{W} \otimes x+b) ,
\end{equation}

\noindent where the tensor product of the weights $(W)$ of individual points $(x)$ are added with the biases $(b)$ vector and $a$ is the non-linear activation function. The convolution operation is fulfilled by three 1-D kernels without striding, the final network by stacking three convolution blocks with the filter sizes (128, 256, 128) in each block.

The integration of LSTM with FCN results in a hyrbid LSTM+FCN classifier, a hybrid approach that combines FCN's temporal feature extraction capabilities with LSTM's sequential data processing strength \cite{karim2017lstm}. This architecture augments the FCN with an LSTM block to process raw time series data, enabling the classifier to capture both spatial and temporal dependencies. Karim et al. utilized a hybrid method consisting of LSTM and FCN (LSTM+FCN) to augment FCN with LSTM for the TSC task with MTS data \cite{karim2019multivariate}. The architecture builds on the FCN's foundation, incorporating three sequential convolutional blocks designed for temporal data analysis, with filters sized 128, 256, and 128, mirroring the setup detailed by Wang et al.~\cite{wang2017time}. Each convolutional segment, comprising a temporal convolutional layer, is refined through batch normalization and employs the ReLU function for activation, leading to robust feature extraction capabilities. The culmination of the convolutional process is marked by global average pooling, ensuring a compressed yet informative feature representation. 

\noindent \emph{InceptionTime:}
InceptionTime is a highly regarded algorithm for MTS classification in DL, frequently considered a benchmark in the field \cite{ismail2020inceptiontime}. Originating from the renowned Inception architecture developed for image recognition via deep convolutions \cite{szegedy2015going}, InceptionTime extends this framework to TSC. It integrates convolutional neural networks (CNNs) with inception modules, utilizing depth-wise separable convolutions, residual connections, and ensemble methods. Each element of InceptionTime is grounded in solid mathematical concepts, aimed at enhancing the classifier's capability to identify and classify intricate temporal patterns. At its core, InceptionTime consists of two distinct residual blocks, each containing multiple inception modules, as illustrated in Fig. \ref{fig:design_dia} (bottom).


Each inception module consists of sliding $z$ filters as the layer transforms the time-series from a MTS with $Z$ dimensions to MTS with $z<<Z$ dimensions. Once the dimensional spaces is reduced and transformed, it is input to sliding filters made of different lengths. The inception blocks make a prediction with even weight, collecting into an ensemble of predictions through the network based on different initializations by:

\begin{equation}
    \hat{y}_{i,c}=\frac{1}{n}\sum_{j=1}^n \sigma_c (X_i,\theta_j) \quad | \quad \forall c \in [1,C] .
\end{equation}

\noindent \emph{RandOm Convolutional KErnel Transform (ROCKET):}
ROCKET is a kernel-based state-of-the-art algorithm for time series classification. Introduced by Dempster et al.~\cite{dempster_rocket_2020}, it represents a significant shift in approach to time series data. ROCKET has the ability to generate a large and diverse set of features, capturing various aspects of the time-series data, which enables highly accurate classification results while utilizing simple linear classifiers. At the heart of ROCKET are its convolutional kernels, which are randomly generated according to specific rules. Each kernel is characterized by its length, dilation, and padding, but unlike in standard CNNs, these attributes are chosen randomly. The intuition is that even though individual kernels might be suboptimal for specific classification tasks, the aggregate effect of many such kernels is a highly discriminative feature set that captures a wide range of patterns in the time series data.

The ROCKET algorithm transforms the time-series dataset through convolutional kernels like traditional convolutional networks. A convolutional kernel $K$, in the context of ROCKET, is defined by its length $l$, dilation rate $d$, and padding strategy. The kernel is applied to a time series $T$ of length $N$ by using the dot product from the kernel and input time series; the convolution operation is:

\begin{equation}
    X_i*\omega = \left(\sum_{j=0}^{l_{kernel}-1}X_{i+(j\times d)} \times\omega_j\right) +b .
\end{equation}

\subsection{Evaluation metrics} \label{Section: mod_inference}
We evaluate classifier performance using the area under the precision-recall curve (AUC-PR), which is particularly informative in the presence of class imbalance \cite{davis2006relationship, brabec2018bad}. In our experiments, the class distribution within folds varied considerably between the two datasets, with positive class ratios ranging from 0\% to 100\%. This was also affected by the TSCV strategy as each strategy experienced a different class distribution, highlighting the need for robust metrics that account for highly skewed class distributions. Extreme cases such as test folds with only a single class (e.g., 0\% or 100\% positive class ratio) were prevalent under the sliding window configuration, especially at higher fold counts and demonstrates how temporal partitioning can amplify class imbalance and distort classifier evaluation if not properly accounted for. Precision $P=\frac{T_p}{T_p+F_p}$ measures the proportion of predicted positive instances that are correct, and recall $R=\frac{T_p}{T_p+F_n}$ quantifies the proportion of actual positives that are successfully identified by the classifier. Then average precision is defined as:

\begin{equation}
\int_0^1 P(R) dR \approx \sum_i^N (R_i - R_{i-1}) P_i ,
\end{equation}

\noindent where $P_n$ and $R_n$ denote the precision and recall at the $i^{th}$ threshold up to the total number of discrete thresholds $N$. This summation computes a weighted mean of precisions, where each weight corresponds to the increase in recall, yielding a scalar value that reflects the classifier's ability to maintain high precision over increasing recall. 

We used the Mann-Whitney U test \cite{mann1947test} and set the significance level to $\alpha=0.05$ to test the null hypothesis that the distribution of two populations have the same distribution. The Mann-Whitney U test is a non-parametric test often used as a test of difference in location between distributions.



	



%% file: 30-Results.tex
 \section{Results} \label{Section: results}
This section present the experimental assessment of the benchmarked classifiers applied to the intermittent fault dataset on different TSCV methods. We emphasize reproducibility and make our code publicly available on GitHub.\footnote{\url{https://github.com/TovNephesh/MTS_CAN}} The proposed methodology was implemented using \textit{Python}, leveraging libraries such as \textit{NumPy} for data manipulation, \textit{SciPy} for statistical analysis, \textit{scikit-learn} for shallow learning, and \textit{TensorFlow} for DL. We evaluate classifier performance using the AUC-PR for imbalanced binary classification. Two TSCV techniques, WF and SW, are employed to simulate streaming and incremental learning conditions, respectively. The analysis is aggregated for all experiments across two distinct CAN datasets representing fault injections in spark-plug and power-wire configurations. We experiment over $K \in \{3,9\}$, with $\omega$ fluctuating and dependant on TSCV strategy and $K$-fold number. $\delta = 150$ was selected based on the size of the available datasets; with the smallest dataset containing 1500 timestamps, $\delta$ represents 10\% of this length, ensuring a meaningful subsequence window while accommodating limited data availability. For statistical rigor, folds with test sets lacking positive or negative instances were excluded, as AUC-PR is undefined in such scenarios. 

We assessed the stationarity of each individual signal in the multivariate time series dataset using both the Augmented Dickey-Fuller (ADF) \cite{dickey1979distribution} and the Kwiatkowski-Phillips-Schmidt-Shin (KPSS) \cite{kwiatkowski1992testing} tests. The ADF test examines the null hypothesis that a series has a unit root (i.e., is non-stationary), while the KPSS test assumes stationarity under the null. Results indicate that at least one series in each fault dataset fails the stationarity condition under both tests, implying that the entire multivariate process is non-stationary. This conclusion is justified by the theoretical requirement that a multivariate time series is stationary only if all component series are individually stationary and their joint covariance structure is time-invariant \cite{lutkepohl2005new, hamilton2020time}. Testing for stationarity is important in this context because non-stationary behavior can affect the temporal consistency of model evaluation under TSCV, potentially biasing performance estimates depending on fold structure.

We conduct five different but complementary analyses. 

First, we investigate whether the choice of TSCV method introduces statistically significant differences in predictive performance (Section \ref{Section:Results_Q1}). Next, we assess the interaction between classifier architecture, comparing shallow classifiers and DL classifiers, and TSCV strategy (Section \ref{Section:Results_Q2}). We then compare classifier-specific performance within each TSCV regime to highlight algorithmic sensitivities to temporal structure (Section \ref{Section:Results_Q3}). Then, we analyze the impact of training-test partition granularity and measure how classifier performance evolves across increasing fold counts (Section \ref{Section:Results_Q4}). Finally, we investigate how classifier performance fluctuates in response to varying degrees of class imbalance across folds, using the area under the AUC-PR versus positive class ratio curve to quantify sensitivity and stability under different TSCV strategies (Section \ref{Section:Results_Q4.2}). In the following results, when we report $p$-values, we assume a significance level $\alpha=0.05$. 


\input{32-TSCV-final}
\input{34-Discussion}

%% file: 32-TSCV-final.tex
\subsection{Performance comparison of TSCV strategy} \label{Section:Results_Q1}
\begin{figure}[!htb]
    \centering
    \includegraphics[width=\columnwidth]{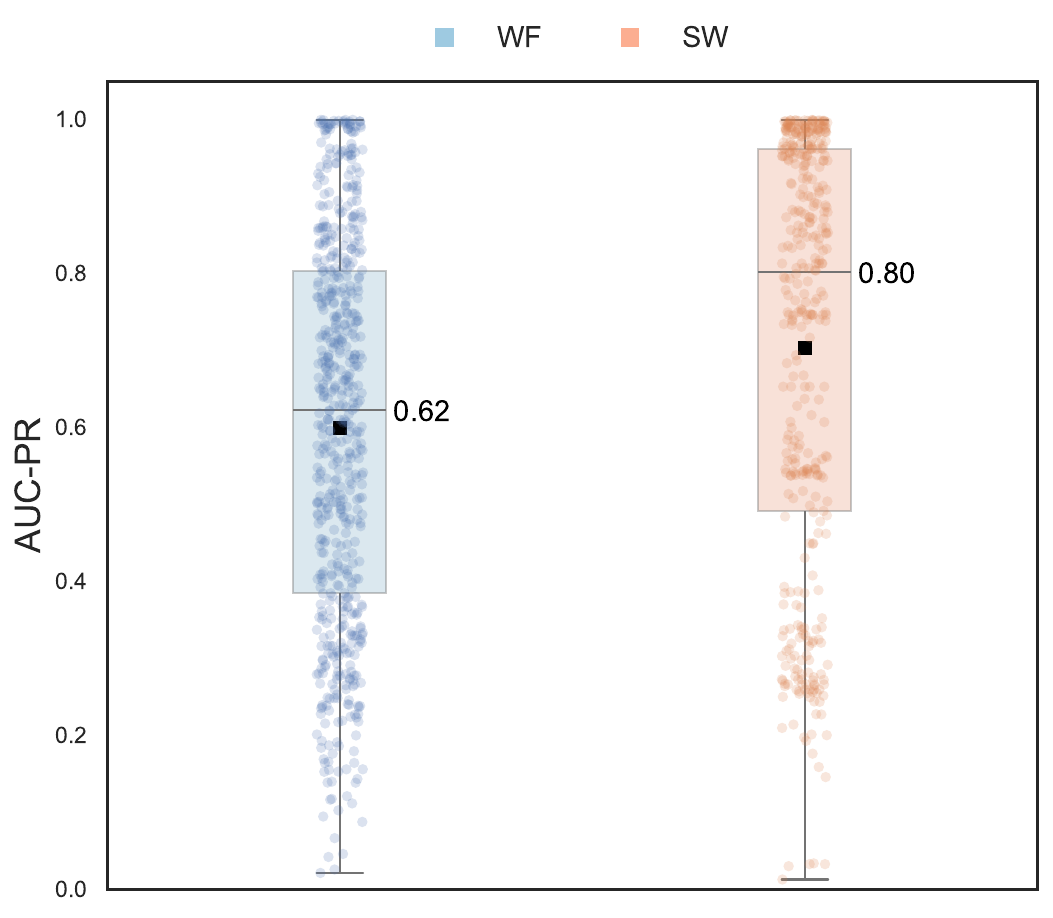}
    \caption{Distribution of AUC-PR scores for all classifiers under WF and SW TSCV strategies. Median is reported numerically and shown with the horizontal line, mean is shown with the black square.}
    \label{fig:wf_sw}
\end{figure}

Temporal dependency matters, as the arrival of observations must be maintained in an experiment involving streaming data \cite{duong2018applying}. While holdout partitioning may preserve subsequnce temporal dependency, it only provides a single classifier training instance on the entire dataset. We begin the evaluation process by testing two TSCV strategies that maintain the sequencital integrity of this dataset while also allowing for retraining. Fig. \ref{fig:wf_sw} separates WF and SW methods and shows the AUC-PR scores via stripplots and distribution of the scores via boxplots. 

Visual inspection of the boxplots and stripplots reveals that SW not only achieves higher median performance but also exhibits higher interquartile range, as evidenced by the box for SW being shifted upward relative to WF. We observe that the middle 50\% of SW scores lies in a higher AUC-PR region compared to WF. In contrast, WF displays a broader spread and greater density of scores below 0.5, suggesting instability in performance across certain folds or classifier configurations. These results collectively highlight that the SW strategy yields both higher and more stable classifier performance, making it a more effective TSCV approach in the context of AUC-PR evaluation. Mann–Whitney U test with one sided alternative hypothesis and  $p = 0.00$, indicate the distributions of AUC-PR scores from the SW method are statistically significantly greater than those from the WF method, providing strong evidence that WF performs worse than SW in terms of AUC-PR. Results of this test provide robust evidence that the selection of TSCV strategy systematically impacts classifier performance distributions. 

\subsection{Interaction between classifier architecture and TSCV} \label{Section:Results_Q2}
\begin{figure}[!htb]
    \centering
    \includegraphics[width=\columnwidth]{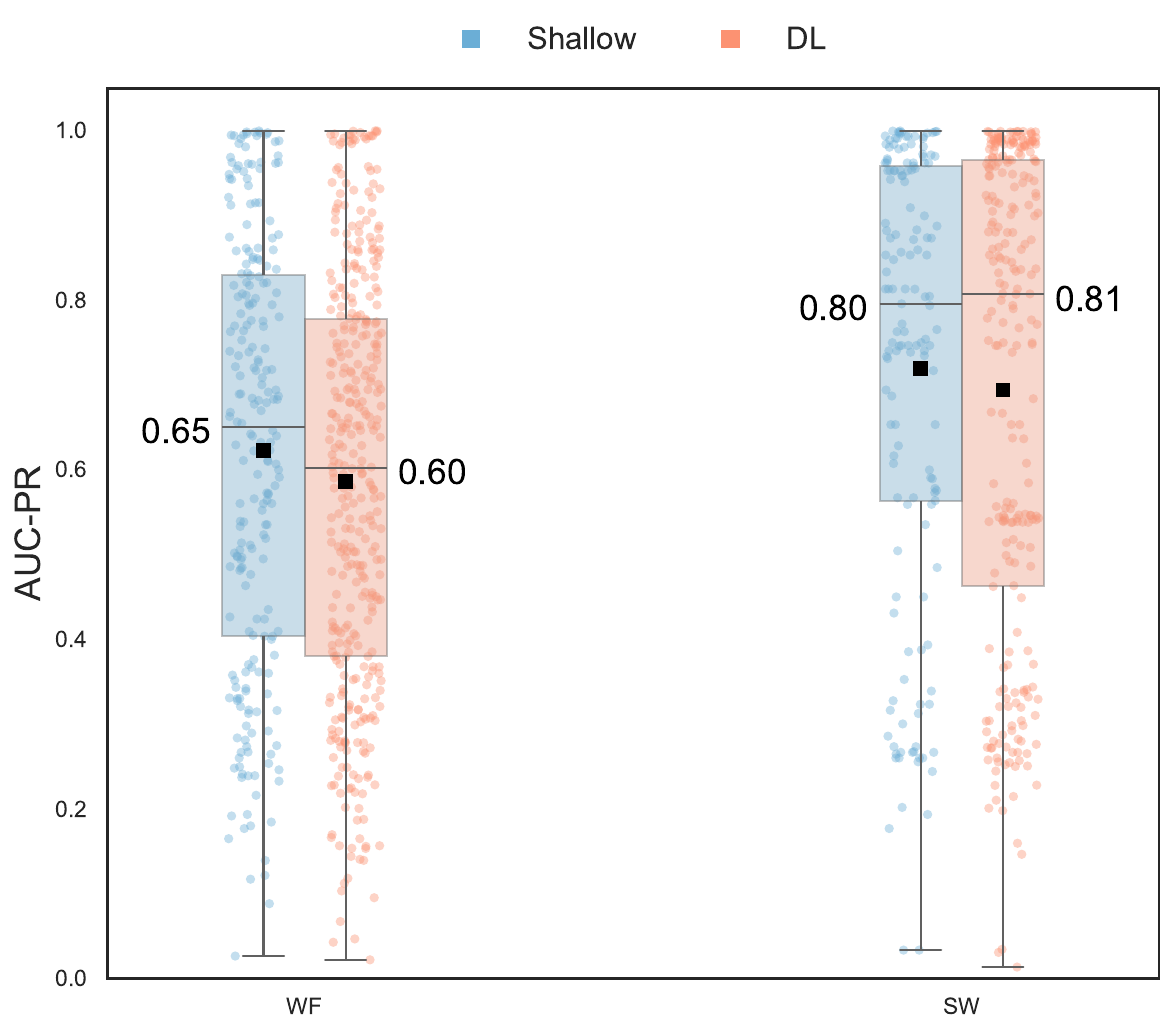}
    \caption{Boxplot and stripplot comparison of AUC-PR distributions for shallow and deep classifiers under WF and SW strategies.}
    \label{fig:wf_sw_shl_dep}
\end{figure}

Next, we compared the AUC-PR scores of shallow and DL classifiers within each TSCV strategy. Fig. \ref{fig:wf_sw_shl_dep} compares boxplots and stripplots in a similar fashion as Fig. \ref{fig:wf_sw}. This analysis centers around comparing shallow and deep across WF and SW. On one hand WF, shallow classifiers exhibit slightly higher median performance (median=0.65) than deep classifiers (median=0.60). On the other hand under SW, this trend reverses, with deep classifiers performing marginally better (median=0.81) than shallow (median=0.80). Although box heights suggest similar interquartile spreads, the shifting medians indicate that classifier architecture interacts with temporal validation to affect classifier behavior, though the differences are modest.

Table \ref{tab:mannu_fam} assess whether classifier architecture depth interacts with the choice of TSCV strategy to significantly affect anomaly detection performance. Using one-sided Mann–Whitney U test, we find no statistically significant difference in AUC-PR performance between shallow and DL classifiers under either WF ($p$-value=0.96) and SW ($p$-value=0.68) strategies. This indicates that, within each TSCV condition, both architecture depths perform comparably in median performance, with no dominant architecture. However, when comparing TSCV methods across the two architectures, both shallow ($p$-value=0.00) and DL ($p$-value=0.00) classifiers demonstrate statistically significant differences between WF and SW. This underscores the sensitivity of classifier performance to the temporal validation design. These findings suggest that while classifier depth does not directly dictate performance in isolation, its interaction with temporal validation strategy is critical and warrants careful consideration during classifier selection subsequence anomaly detection in time-ordered data. 

\begin{table}[!ht]
\centering
\begin{tabular}{ c | c | c }
\hline
Comparison & Condition & $p$-value \\
\hline
Shallow vs. DL & WF & 0.96 \\
Shallow vs. DL & SW & 0.68 \\
WF vs. SW & DL & \textbf{0.00} \\
WF vs. SW & Shallow & \textbf{0.00}\\
\hline
\end{tabular}
\caption{Results of Mann–Whitney U tests comparing AUC-PR scores across classifier architecture (Shallow vs. DL) and TSCV strategies (WF vs. SW).}
\label{tab:mannu_fam}
\end{table}

\subsection{Comparison of classifiers} \label{Section:Results_Q3}
\begin{figure*}[!ht]
    \centering
    \includegraphics[width=\textwidth]{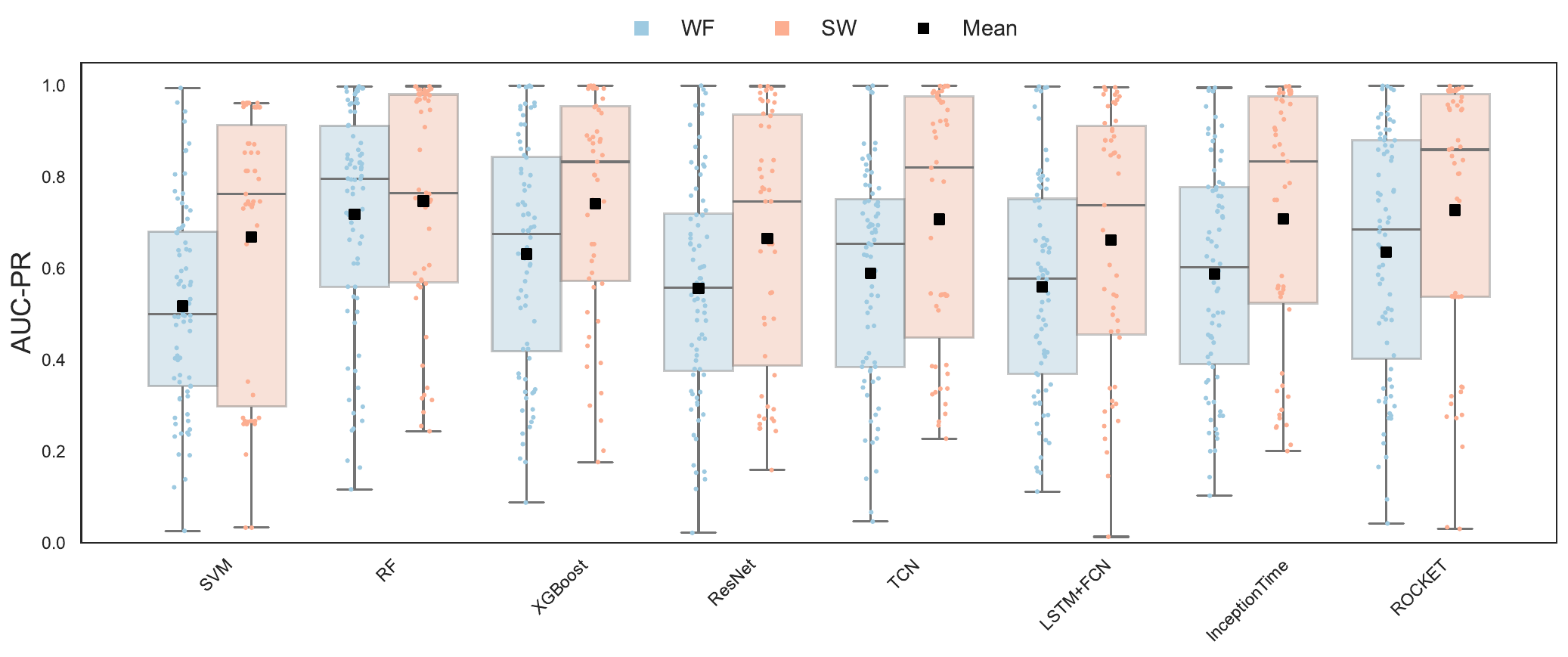}
    \caption{Boxplot and stripplot comparison of AUC-PR scores across classifiers under two TSCV strategies (WF and SW). Each box represents the performance distribution of a specific classifier across all $K$-folds. Black horizontal lines denote median and black squares denote the mean AUC-PR.}
    \label{fig:cls_tscv}
\end{figure*}

A critical dimension of this study is understanding how individual state of the art classifiers respond to different TSCV strategies. Presented here is the exploration of the classifier-level TSCV comparison. Fig. \ref{fig:cls_tscv} illustrates the distribution of AUC-PR scores for each classifier across WF and SW TSCV strategy. Interquartile ranges tend to be shorter and shifted down for classifiers operated with the WF strategy suggesting less variability with lower AUC-PR scores. Generally, all classifiers exhibit higher AUC-PR scores when operating with the SW method. Most classifiers operated with SW whiskers indicate AUC-PR score distributions are skewed potentially to left. 

Table \ref{Table: RQ3} highlights the medians of AUC-PR scores and $p$-value results of the one-sided Mann-Whitney U test that takes each classifier and compares them across the TSCV strategies. All classifiers benefit from an increase in median AUC-PR score under the SW strategy compared to the WF strategy. Based on $p$-values, most classifiers demonstrated statistically significant improvements under the SW strategy compared to WF, suggesting most classifiers benefit from the localized temporal context preserved in SW evaluations. Notably, XGBoost exhibited a substantial median gain from 0.67 (WF) to 0.83 (SW) and a $p$-value of 0.01. Similarly, all DL classifiers exhibited significant improvements under the SW regime compared to WF. These findings highlight a consistent pattern: more expressive classifiers are better able to capitalize on the temporal stationarity that SW cross-validation maintains by evaluating over temporally overlapping windows.

Among shallow learners, SVM displayed a particularly pronounced shift when moving from WF to SW in performance with a median AUC-PR from 0.50 to 0.76 and $p$-value = 0.00, suggesting a heightened sensitivity to the increased sample diversity or temporal ordering introduced by SW. RF is the only classifier that showed only a marginal median AUC-PR score decrease from 0.80 to 0.77 and failed to reach statistical significance with a $p$-value = 0.35. This suggests that ensemble tree-based methods may already exhibit robustness to the temporal decomposition of training data. Taken together, these results suggest that SW may offer a more reliable evaluation framework for detecting subtle temporal TSCV, particularly for classifiers with high capacity and temporal encoding ability. However, the varying degree of sensitivity across classifiers also underscores the importance of aligning classifier complexity and validation strategy on MTS subsequence anomaly detection evaluation.

\begin{table}[htbp]
  \centering
    \begin{tabular}{p{.09cm} | p{1.9cm} |cc|c}
    \hline
    & Classifier & {$M_{\text{WF}}$} & {$M_{\text{SW}}$} & {$p$-value} \\
    \hline
    \multirow{3}{*}{\rotatebox[origin=c]{90}{Shallow}}
    & SVM   & 0.50  & 0.76  & \textbf{0.00} \\
    & RF    & 0.80  & 0.77  & 0.18 \\
    & XGBoost & 0.67  & 0.83  & \textbf{0.00} \\
    \hline
    \multirow{5}{*}{\rotatebox[origin=c]{90}{DL}}
    & ResNet & 0.56  & 0.75  & \textbf{0.01} \\
    & TCN   & 0.65  & 0.82  & \textbf{0.00} \\
    & LSTM+FCN & 0.58  & 0.74  & \textbf{0.01} \\
    & InceptionTime & 0.60  & 0.83  & \textbf{0.00} \\
    & ROCKET & 0.69  & 0.86  & \textbf{0.00} \\
    \hline
    \end{tabular}
    \caption{Classifier-wise comparison of median ($M$) AUC-PR scores under WF and SW TSCV strategies. Included are median performance across all $K$-folds for each classifier, along with $p$-values from Mann–Whitney U tests assessing statistical differences between the two TSCV strategies.}
    \label{Table: RQ3}
\end{table}%

\subsection{Training opportunity investigation} \label{Section:Results_Q4}
\begin{figure*}[!htb]
    \centering
    \includegraphics[width=\textwidth]{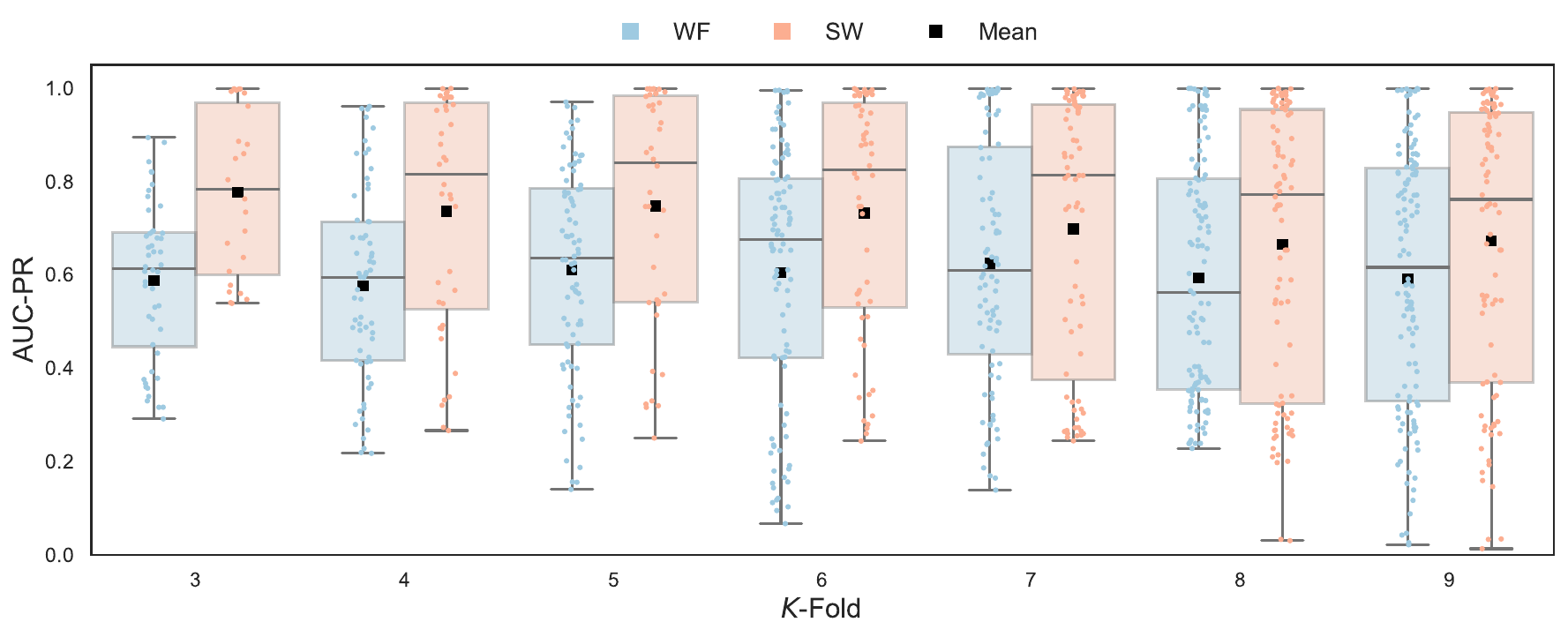}
    \caption{Boxplot and stripplot of distributions of AUC-PR scores across varying values of $K$-folds for WF and SW TSCV strategies. Black horizontal lines denote median and black squares denote the mean AUC-PR.}
    \label{fig:sw_rank_sp}
\end{figure*}

Next, we investigate the effect that the number of training opportunities has an effect on the classifiers. The underlying hypothesis is that as the number of $K$-folds increases, classifiers benefit from a greater volume of training data per fold, which can positively affect generalization during testing. Fig. \ref{fig:sw_rank_sp} illustrates the distribution of AUC-PR scores across increasing values of $K$-folds for both WF and SW TSCV strategies, providing insight into how training opportunity volume influences classifier performance. SW consistently demonstrates higher median AUC-PR values, suggesting SW better captures temporal dependencies or leverages more stable evaluation intervals. WF distributions appear to be more symmetrical while SW distributions are skewed left. 

Table \ref{Table: kfold} showcases median classifier AUC-PR scores for WF and SW at the $K$-fold level and one-sided Mann-Whitney U tests for comparing WF and SW at each $K$-fold. A consistent performance advantage is observed for SW across $K=3$ through $K=6$, with statistically significant differences ($p$-value = 0.00). This suggests that SW yields more favorable conditions for classifier learning with less training opportunity. This performance gain is likely due to the overlapping nature of SW folds, which offer denser temporal coverage and more representative exposure to intermittent fault signatures during training, allowing classifiers to generalize more effectively.

As $K$ increases beyond 6, the magnitude of improvement for SW diminishes and the statistical significance weakens. At $K=7$, although the median AUC-PR for SW (0.81) remains higher than WF (0.61), the $p$-value of 0.07 indicates that this difference is no longer statistically significant. However, at $K=8$ and $K=9$, the $p$-values of 0.04 and 0.00 respectively indicate that the performance benefit of SW reemerges with statistical significance. These results suggest that while the advantage of SW may temporarily diminish around moderate $K$, it can persist or resurface at higher folds, potentially due to SW’s ability to maintain sufficient training context even when test segments become smaller. Overall, the analysis highlights that the advantage of SW is most prominent at lower to moderate values of $K$, but becomes less impactful in high-fold cross-validation settings.

\begin{table}[htbp]
  \centering
    \begin{tabular}{r|rr|r}
    \hline
    \multicolumn{1}{l|}{$K$-fold} & \multicolumn{1}{l}{$M_{\text{WF}}$} & \multicolumn{1}{l}{$M_{\text{SW}}$} & \multicolumn{1}{|l}{$p$-value} \\
    \hline
    3 & 0.61 & 0.78 & \textbf{0.00} \\
    4 & 0.60 & 0.82 & \textbf{0.00} \\
    5 & 0.64 & 0.84 & \textbf{0.00} \\
    6 & 0.68 & 0.83 & \textbf{0.00 }\\
    7 & 0.61 & 0.81 & 0.07 \\
    8 & 0.56 & 0.77 & \textbf{0.04} \\
    9 & 0.62 & 0.76 & \textbf{0.00} \\
    \hline
    \end{tabular}
    \caption{$K$-fold-based summary of the AUC-PR score medians across WF and SW TSCV and $p$-value results from statistical testing of median differences between WF and SW across $K$-folds.}
    \label{Table: kfold}
\end{table}%

\subsection{Quantifying fold-level imbalance effects} \label{Section:Results_Q4.2}
While Section \ref{Section:Results_Q4} examined how varying training opportunities through different $K$-folds impacts overall performance, we now turn to a more granular analysis, assessing how classifier performance fluctuates in response to varying class imbalance across individual test folds. To quantify how sensitively and consistently a classifier’s performance responds to changes in class imbalance during time-sequenced evaluations we calculate AUC of AUC-PR scores and positive test ratio. The test positive ratio reflects the proportion of positive samples within the test set at each fold, allowing us to assess how classifier performance scales with varying levels of class imbalance for each TSCV strategy. Here, we evaluate all AUC-PR scores at each fold inside a particular $K$-fold across each TSCV strategy. We have $K \in \{3, 4, \dots, 9\}$ partitioning scheme, where folds with only one class represented skipped as AUC-PR can't be calculated in that scenario. For a given classifier $C$ and fold setting $K$, let $\mathcal{F}_{C,K} = \{(r_k, s_k)\}_{k=1}^K$ be the set of fold-level evaluation pairs, where $r_k$ is the positive test ratio and $s_k$ is the corresponding AUC-PR score on fold $k$. Assuming the values are sorted by $r_k$, the AUC is computed using the trapezoidal rule as:
\begin{equation}
\text{AUC}(\mathcal{C}) = \int_{r_{k \min}}^{r_{k \max}} s(r)\,dr \approx \sum_{k=1}^{K-1} \frac{s_k + s_{k+1}}{2} (r_{k+1} - r_k),
\end{equation}

\noindent where $s_k$ is the AUC-PR on  $\boldsymbol{S}_{test}^{k}$ and $r_k$ is the test positive ratio associated with that fold.

Table \ref{tab:auc_classifier_summary} highlights the AUC and quantifies the sensitivity of classifier performance to variations in class imbalance across test folds, thus serving as a proxy for generalizability under real-world sampling conditions. We observe that across all classifiers, the SW method tends to produce higher median AUC values than the WF method, particularly evident in classifiers such as XGBoost (0.43 vs. 0.37), InceptionTime (0.43 vs. 0.36), and TCN (0.44 vs. 0.35). The only cases in which a classifier produced higher median values for WF compared to SW was RF ($0.44$ vs. $0.40$) and LSTM+FCN ($0.40$ vs. $0.38$). These differences indicate that SW enables more stable performance as the proportion of positive examples varies in test data, likely due to its more gradual evolution of training and test sets across time. Additionally, the lower standard deviations under SW—especially for DL such as LSTM+FCN and ResNet—suggest reduced performance volatility across folds.

We notice that shallow classifiers exhibit slightly higher AUC values in absolute terms compared to most DL, indicating their robustness to changing class balance during evaluation. However, their higher standard deviations under WF, particularly RF (0.20) and XGBoost (0.15), suggest that their stability is more sensitive to the choice of TSCV strategy. The DL classifiers show more consistent AUC behavior across both WF and SW methods, implying that architectural features such as temporal filters or recurrence may better capture patterns that generalize across imbalance scenarios. Collectively, the results reinforce the utility of SW for stabilizing classifier performance under class ratio variability and highlight differences in temporal robustness between shallow and deep architectures.

\begin{table}[!ht]
\centering
\begin{tabular}{p{.08cm} | p{1.9cm} | c c | c c}
\hline
& \textbf{Classifier} & $M_{\text{WF}}$ & $\sigma_{\text{WF}}$ & $M_{\text{SW}}$ & $\sigma_{\text{SW}}$ \\
\hline
\multirow{3}{*}{\rotatebox[origin=c]{90}{Shallow}}
& SVM           & 0.37 & 0.14 & 0.40 & 0.09 \\
& RF            & 0.44 & 0.20 & 0.40 & 0.12 \\
& XGBoost       & 0.37 & 0.15 & 0.43 & 0.12 \\
\hline
\multirow{5}{*}{\rotatebox[origin=c]{90}{DL}}
& ResNet        & 0.35 & 0.14 & 0.37 & 0.10 \\
& TCN           & 0.35 & 0.17 & 0.44 & 0.10 \\
& LSTM+FCN      & 0.40 & 0.15 & 0.38 & 0.09 \\
& InceptionTime & 0.36 & 0.15 & 0.43 & 0.11 \\
& ROCKET        & 0.38 & 0.18 & 0.41 & 0.11 \\
\hline
\end{tabular}
\caption{Classifier-based summary of the AUC between AUC-PR and test positive class ratio for each TSCV method (WF and SW). The table reports the median ($M$) and standard deviation ($\sigma$) of AUC values across $K$-folds.}
\label{tab:auc_classifier_summary}
\end{table}

Finally, we evaluate how classifier performance, in terms of precision-recall tradeoff, varies as a function of the test set's positive class ratio across folds. By computing the AUC between AUC-PR scores and test positive ratios for each fold, this metric quantifies how sensitive classifier performance responds to class imbalance under both WF and SW strategies. Table \ref{tab:auc_kfold_summary} shows that WF method exhibits a steep increase in medidan AUC values from $K=3$ to $K=7$, peaking at $M_{WF}$ of 0.57 at $K=7$, and tapering off at $K=9$. In contrast, the SW approach displays a more immediate and consistent plateau, with median AUC values stabilizing between $M_{SW}=0.40-0.47$ at $k\geq4$. The standard deviation for WF is slightly larger than that of SW especially at higher $K$-folds (except $K=8$), indicating greater variability in classifier responses under WF. This suggests that while classifiers benefit with WF from more training sets, it also introduces more pronounced shifts in test class distributions across folds. The more consistent performance of SW may be attributed to its fixed windowing mechanism, which experiences different class distributions compared to WF over time. These results underscore that TSCV strategy and the number of folds both influence how robustly classifiers respond to binary class distributions in temporal scenarios.

\begin{table}[!ht]
\centering
\begin{tabular}{c| c c | c c}
\hline
$K$-fold & $M_{\text{WF}}$ & $\sigma_{\text{WF}}$ & $M_{\text{SW}}$ & $\sigma_{\text{SW}}$ \\
\hline
3 & 0.14 & 0.01 & 0.17 & 0.01 \\
4 & 0.18 & 0.02 & 0.40 & 0.04 \\
5 & 0.32 & 0.04 & 0.42 & 0.02 \\
6 & 0.41 & 0.07 & 0.40 & 0.04 \\
7 & 0.57 & 0.05 & 0.40 & 0.02 \\
8 & 0.37 & 0.03 & 0.47 & 0.06 \\
9 & 0.47 & 0.07 & 0.45 & 0.04 \\
\hline
\end{tabular}
\caption{$K$-fold-based summary of the AUC between AUC-PR and test positive class ratio for each TSCV method (WF and SW). The table reports the median ($M$) and standard deviation ($\sigma$) of AUC values across classifiers.}
\label{tab:auc_kfold_summary}
\end{table}

%% file: 34-Discussion.tex
\section{Discussion}
This study provides an in-depth evaluation of the impacts of TSCV strategy on shallow and DL classifiers applied to MTS subsequence anomaly detection. We focus on WF and SW TSCV strategies. The results indicate that the best configuration is utilized with a SW TSCV strategy. Tree-based shallow learners appear to have a slightly improved fault detection ability compared to other classifiers as XGBoost and RF slightly outperformed the DL classifiers and SVM across AUC-PR metrics. These findings highlight the suitability of shallow learning classifiers for handling temporal dependencies in streaming fault detection environments and are particularly advantageous due to their scalability. Interestingly, the RF classifier traverses across TSCV strategy without statistical significant differences. Despite their complexity and capability to capture intricate patterns, DL classifiers don't generalize better compared to shallow learning classifiers, particularly with fluctuating results across different folds and methods. With the exception of RF, classifiers are significantly impacted by the TSCV strategy, suggesting they might be sensitive to the specific data partitioning methods and require further tuning to fully leverage their architecture's promise.

\subsection{General observations}
Our results demonstrate that the SW TSCV method provided the most consistent performance for both shallow and DL classifiers. SW maintained the highest AUC-PR scores for both classifier types, with XGBoost and RF achieving the highest scores. The SW approach’s ability to preserve temporal dependencies likely contributed to this success, ensuring that classifiers were evaluated on temporally consistent data. The consistently strong performance of RF in AUC-PR (esspecially under varying test class distributions) underscores RFs robustness in capturing fault and non-fault zones. RF excelled across all TSCV methods, further reinforcing the efficacy of tree-based methods in structured and time-dependent data. These results imply that shallow learning classifiers may be better suited to tasks where fault detection must occur in real time, with minimal computational overhead.

\subsection{Insights of TSCV strategies}

The comparative analysis of TSCV strategies reveals that SW consistently outperforms WF in terms of both classification and stability across a variety of classifiers. This advantage is particularly evident in lower to moderate $K$-fold settings, where SW's overlapping evaluation windows preserve local temporal dynamics and offer denser, more representative training exposures. The statistical superiority of SW is supported by consistently higher mean AUC-PR values and significantly lower performance volatility in DL classifiers, indicating that SW better accommodates the inherent non-stationarity of intermittent faults. In contrast, WF exhibits greater variability and skewness in performance distributions, suggesting heightened sensitivity to fold-specific class imbalance and limited adaptability to abrupt shifts in fault patterns. Furthermore, the AUC analyses between AUC-PR and test positive ratio demonstrate that SW maintains more robust classifier behavior under varying imbalance conditions, reinforcing its value for real-world deployment where consistent performance under drift and sampling variability is critical. Overall, these insights underscore the importance of selecting TSCV methods that align with the temporal and structural characteristics of streaming diagnostic data.

Cerqueira et al. \cite{cerqueira2020evaluating} similarly emphasize the importance of preserving temporal structure in cross-validation for accurate performance estimation. They found that growing windows (akin to WF) was the best approach for univariate non-stationary TS data. Our findings do not support this claim as TSCV is applied to non-stationary MTS with class imbalances. Our findings suggest that overlap-based methods like SW better accommodate evolving fault dynamics and should be prioritized in temporal classification scenarios.

\subsection{Findings from training opportunities}
With testing the TSCV strategies, we are able to evaluate how well these classifiers generalize and with comparing results from different $K$-folds, we discovered the optimized amount of training size for best results. The order and continuity of the data are crucial. Events in time series data, especially CAN data, can be influenced by patterns that evolve gradually over time. By not overlapping the training and test sets, the WF strategy appears to struggle with the most recent test data during future training rounds. Therefore, the WF method experiences a loss of recent context in the data and discontinuity since the training data only includes information up to a certain point and the test set contains newer unseen data. SW also makes more efficient use of the data because each window contributes to both training and testing. The overlap ensures that the classifier sees multiple perspectives on similar time periods. SW is not perfect, since the method reuses data across multiple windows, there is potentially a risk of overfitting.

Tables \ref{Table: kfold} and \ref{tab:auc_kfold_summary} reveal the intricacies how fold selections impact classifiers.  

\subsection{Limitations}
While the findings are promising, our work exhibits the following limitations.

\noindent \textbf{Dataset size:} The two MTS datasets used for this study consists of 2,490 and 1500 timestamps, each timestamp representing 0.1 seconds of data. These relatively small dataset sizes may have hindered the performance of the DL classifiers, which typically require larger datasets to fully optimize their learning capacity. The dataset’s limited size may also not reflect the full range of potential fault conditions in real-world CAN systems, limiting the generalizability of these findings.

\noindent \textbf{Class imbalance:} An additional challenge noted in the experiment is the class imbalance present in certain folds of the data. In intermittent fault detection, some folds contained few or no examples of fault events, potentially creating bias (folds with one class representation are skipped during evaluation) since these folds could not be utilized for calculating AUC-PR. This class imbalance could significantly affect the classifiers' ability to generalize well across different folds and warrants further study into methods for balancing the classes in time series data. It also limits comparability of the two TSCV methods as class imbalance is different for WF and SW.

\noindent \textbf{Use of default classifier configurations:} To ensure consistency and reduce experimental complexity, default architectures and hyperparameters (fixed filter sizes, 100 estimators, learning rate=0.01, etc.) were used across classifiers following the spirit of previous studies \cite{hespeler2021online, hespeler2022deep, hespeler2024deep}. While this creates uniformity for baseline comparability, it may have constrained the potential of certain classifiers, particularly the DL classifiers, which typically benefit from extensive hyperparameter tuning and architecture optimization to achieve their full potential. 

\noindent \textbf{Use of simplified window:} The scope of this study did not extend to analyzing the effects of optimized SW parameters (like window length and offset) on classifier performance evaluation. Instead, a fixed-parameter SW configuration was employed to ensure consistent and controlled comparisons with the WF strategy. While this design promotes methodological comparability across TSCV techniques, it may underutilize the full potential of the SW approach. 

\subsection{Broader implications}
These findings have important implications for the field of subsequence anomaly detection, particularly in streaming environments. The success of traditional ML classifiers, especially when coupled with SW, suggests that simpler classifiers may outperform more complex DL architectures in environments where dataset size is constrained or where computational efficiency is critical. Moreover, the results highlight the importance of temporal evaluation methods, such as the SW, which preserve the sequential integrity of data and lead to more reliable performance metrics. This is particularly relevant for real-time fault detection, where it is essential to accurately predict faults as new data arrives without retraining the classifier from scratch.

%% file: 50-Conclusion.tex
\section{Conclusion}

This study investigated the impacts of TSCV strategy through state-of-the-art shallow ML and DL for subsequence anomaly detection in MTS. We focus on a case study for detecting faults in CAN. By employing WF and SW strategies across a variety of $K$-fold arrangements, this study evaluates the performance of eight classifiers using AUC-PR scores. 

The experimental results reveal that the choice of TSCV strategy substantially influences classification performance. SW consistently outperformed WF in both median AUC-PR and performance stability (lower variance across classifiers and folds), especially for DL classifiers that benefit from localized temporal continuity. SW was most effective at lower to moderate fold counts, where overlapping evaluation intervals preserved fault signatures and improved classifier generalization. The RF classifier exhibited stable performance across both TSCV strategies, with no statistically significant differences detected, suggesting tree-based robustness to variations in temporal validation design.

Future works could include the use of an ensemble approach that could combine the strengths of multiple classifiers to enhance fault detection reliability. The issue of class imbalance within certain folds could provide useful insight for handling intermittent faults. Future studies should utilize larger, more diverse datasets that better represent real-world operating conditions. This would allow DL classifiers to fully utilize their capacity for complex pattern recognition and may improve their generalizability across different fault types. Conducting a thorough hyperparameter tuning for both shallow ML and DL classifiers could yield better performance. For DL classifiers, in particular, tuning learning rates, batch sizes, and other architectural parameters could allow them to better handle temporal dependencies and variability in data.

%% file: 91-Acknowledgement.tex
\section*{Acknowledgement}


This research was sponsored in part by Oak Ridge National Laboratory’s (ORNL’s) Laboratory Directed Research and Development program and by the DOE. There was no additional external funding received for this study. This research used birthright cloud resources of the Compute and Data Environment for Science (CADES) at the Oak Ridge National Laboratory, which is supported by the Office of Science of the U.S. Department of Energy under Contract No. DE-AC05-00OR22725. The funders had no role in study design, data collection and analysis, decision to publish, or preparation of this manuscript.